\documentclass[sigconf,natbib=true]{acmart}

\AtBeginDocument{%
  \providecommand\BibTeX{{%
    \normalfont B\kern-0.5em{\scshape i\kern-0.25em b}\kern-0.8em\TeX}}}

\copyrightyear{2022}
\acmYear{2022}
\setcopyright{acmcopyright}\acmConference[CIKM '22]{Proceedings of the 31st ACM
International Conference on Information and Knowledge Management}{October
17--21, 2022}{Atlanta, GA, USA}
\acmBooktitle{Proceedings of the 31st ACM International Conference on Information
and Knowledge Management (CIKM '22), October 17--21, 2022, Atlanta, GA, USA}
\acmPrice{15.00}
\acmDOI{10.1145/3511808.3557260}
\acmISBN{978-1-4503-9236-5/22/10}

\usepackage{subcaption}
\usepackage{placeins}
\usepackage{caption}
\usepackage{comment}
\usepackage{todonotes}
\presetkeys
    {todonotes}
    {inline,backgroundcolor=yellow}{}

\newcommand\koby[1]{\textcolor{blue}{[KB: #1}]}

\usepackage{stfloats}
\usepackage{xspace}

\newcommand{\cftolabel}{CF2Label}\xspace
\newcommand{\mtlreconstruct}{\texttt{MTL-reconstruct}\xspace}
\newcommand{\imageonly}{\texttt{Image-only}\xspace}
\newcommand{\contrastiveloss}{\texttt{Contrastive-loss}\xspace}
\newcommand{\finetuning}{\texttt{Sequential}\xspace}

\begin{document}


\title{Collaborative Image Understanding}

\author{Koby Bibas}
\email{kobybibas@gmail.com}
\affiliation{%
  \institution{Meta}
  \city{Tel Aviv}
  \country{Israel}
}
\author{Oren Sar Shalom}
\email{oren.sarshalom@gmail.com}
\affiliation{%
  \institution{Amazon}
  \city{Tel Aviv}
  \country{Israel}
}
\author{Dietmar Jannach}
\email{dietmar.jannach@aau.at}
\affiliation{%
  \institution{University of Klagenfurt}
  \city{Klagenfurt}
  \country{Austria}
}

\begin{abstract}
Automatically understanding the contents of an image is a highly relevant problem in practice. In e-commerce and social media settings, for example, a common problem is to automatically categorize user-provided pictures. Nowadays, a standard approach is to fine-tune pre-trained image models with application-specific data. Besides images, organizations however often also collect \emph{collaborative} signals in the context of their application, in particular how users interacted with the provided online content, e.g., in forms of viewing, rating, or tagging. Such signals are commonly used for item recommendation, typically by deriving \emph{latent} user and item representations from the data. In this work, we show that such collaborative information can be leveraged to improve the classification process
of new images. Specifically, we propose a multitask learning framework, where the auxiliary task is to reconstruct collaborative latent item representations. A series of experiments on datasets from e-commerce and social media demonstrates that considering collaborative signals helps to significantly improve the performance of the main task of image classification by up to 9.1\%.
\end{abstract}



\maketitle


\section{Introduction}
Image understanding can be described as the process of automatically analyzing unstructured image data in order to extract knowledge for specific tasks. Automatically assigning images to pre-defined categories (i.e., image categorization) is one important form of image understanding and an area in which we observed substantial progress in recent years, largely due to innovations in deep learning \cite{imagenet_cvpr09, krizhevsky2012imagenet,bibas2021learning}. Nowadays, one standard way of developing an image categorization solution for a specific task or application is to rely on  pre-trained image models 
and to fine-tune them with application-specific data for the particular problem at hand.

In a specific application setting, additionally to the raw image information (pixels), other types of information may be available as well. In particular, providers of e-commerce or social media sites nowadays often have access to  \emph{collaborative} information that is collected in the context of an application. Such sites usually often record how users interact with the provided online content, e.g., in the form of viewing, rating, commenting, tagging, or resharing. 

Such collaborative signals can be used for different purposes and in particular for personalization. Most importantly, these signals have been successfully used for many years by online retailers and service providers to build \emph{collaborative filtering} (CF) recommender systems \cite{ResnickGrouplens1994}. 
Today's most effective CF systems are based on deriving \emph{latent} representations (\emph{embeddings}) of users and items from a given user-item interaction matrix through matrix factorization \cite{hu2008collaborative} or deep learning techniques \cite{liang2018variational, hidasi2017dlrs}.

The latent representations that are derived from collaborative information cannot be directly interpreted. However, it seems intuitive to assume that the \emph{item representations} may encode some information about the relevant categories in a domain.

In this work, we explore an approach which we call \emph{collaborative image understanding.} Specifically, the idea is to leverage the collaborative information as additional supervision for the \emph{training} phase of image understanding models. By incorporating this collaborative information only during training, the model can support also new images for which collaborative information is not available. Such problems for example arise when online platforms have to automatically process images as soon as they are uploaded, before they gain any collaborative information. Any application that combines images and interaction data suits to this setting, where applications include e-commerce, social media and online content.


The main contributions of our work are as follows.
\begin{enumerate}
    \item We introduce a new signal (i.e., collaborative information) to support the training procedure of automated image understanding processes.
   \item We propose a general multitask-learning (MTL) \cite{caruana1997multitask} framework to incorporate collaborative information in the form of latent item representations into existing single-task content understanding models.
   \item We explore several alternative approaches to combining collaborative information in the training procedure of image classification models.
   \item We show that our approach is also particularly effective in typical real-world situations when labels are missing for a fraction of the training images.
\end{enumerate}
Our MTL approach, where the auxiliary task is to reconstruct collaborative latent item representations, is highly promising.
An in-depth empirical evaluation on the Pinterest~\cite{zhong2015predicting}, MovieLens~\cite{harper2015movielens}, Amazon Clothing and Amazon Toys~\cite{mcauley2015image} datasets revealed that the classification performance can be improved by 5.2\%, 2.6\%, 9.1\%, and 3.0\% respectively.

The rest of this paper is organized as follows. In Section~\ref{sec:background} we provide the necessary background and review earlier work. Section~\ref{sec:approach} details the proposed approach; thorough experiments are described in Section~\ref{sec:experiments}. Section~\ref{sec:discussion} concludes the paper.

\section{Background and Related Work}\label{sec:background}
In this section, we first briefly provide some relevant background in three research areas: image classification, collaborative filtering and multi-task Learning (Section~\ref{subsection:background}). Then, we discuss earlier works, which use side information or collaborative information in combination with image data (Section~\ref{subsection:related-work}).

\subsection{Background}
\label{subsection:background}
\subsubsection{Image Classification with Neural Networks}
Classifying images to predefined categories based on their content is an important task in the areas of computer vision and information extraction. The training input for this learning problem is a set $\{(x_i, y_i)\}$, where $x_i$ is an image of an entity $i$ and $y_i$ is the corresponding category. The ImageNet database \cite{imagenet_cvpr09} for example, contains millions of hand-labeled images, which are assigned to thousands of categories.
Usually models trained on one domain do not perform well on other domains and labels from the target domain are required \cite{zhou2021domain}. As manual annotation is expensive and time consuming, label paucity is a ubiquitous problem.

In the last ten years, substantial progress was made in the field due to advances in deep learning technology and in particular through the use of convolutional neural networks (CNN)
\cite{krizhevsky2012imagenet,vggnet2014,googlenet2015}. Continuously new architectures were proposed over the years, including the ResNet \cite{he2016deep}, MobileNet \cite{howard2017mobilenets} and the recent RegNet \cite{radosavovic2020designing} approaches that we use in the experiments in this paper.

Given a particular application problem, a common approach to mitigate the paucity of labeled data is to rely on existing models that are pre-trained on a large corpus like ImageNet, and to then fine-tune them with the data of the particular application. Such an approach is also followed in our work. However, in addition to the application-specific images and labels, we also consider collaborative information to further improve classification performance.


\subsubsection{Collaborative Filtering and Latent Factor Models}
Collaborative filtering recommender systems are commonly based on a matrix that contains recorded interactions between users and items \cite{ResnickGrouplens1994}, with the typical goal of predicting the missing entries of the matrix and/or ranking the items for each user in terms of their estimated relevance.
Matrix factorization (MF) techniques prove to be highly effective for this task and are based on representing users and items in a lower-dimensional latent space. While early methods were based on Singular Value Decomposition \cite{Billsus1998}, later approaches used various types of machine learning models to compute the latent factors, e.g., \cite{hu2008collaborative}. Note that while various deep learning approaches were proposed for the task in recent years, latent factor models remain highly competitive and relevant today \cite{Rendle2020}.

Technically, MF methods project users and items into a shared latent space, where each user and each item is represented through a real-valued vector. A user vector indicates to what extent the corresponding user prefers each of the latent traits; likewise, an item vector represents to what extent the corresponding item possesses them \cite{koren2009matrix}. The traits are called latent because their \emph{meaning} is unknown. Yet, we know that latent factor models are highly effective for recommendations and furthermore, that user vectors hold information on the demographics of the users \cite{resheff2018privacy}. Therefore, we hypothesize that item vectors encode certain types of relevant information about the items---like category or brand---which we try to leverage in our work for improved image understanding.

\subsubsection{Multi-Task Learning}
MTL is a machine learning approach which optimizes multiple learning tasks \emph{in parallel} based on a shared representation for all tasks. The promise of MTL is that if the tasks share significant commonalities, then this approach may lead to improved performance on each individual task, compared to training task-specific models separately \cite{caruana1997multitask}.
The rationale behind MTL and its success lies in the potential of improved generalization by using the information contained in the signals of related tasks. Thus, each individual task can ``help''  other tasks, often in order to also prevent overfitting for one  particular task.  
Concretely, this is achieved by having some shared parameters for all tasks. That way, during training time each task directly affects the other ones.

Various successful applications of MTL are reported in the literature, both for computer vision problems and other areas, see \citet{Zhang2017MTL} for a survey. However, using MTL to combine collaborative signals and image information for higher categorization accuracy has not been explored in the literature before. As our experimental evaluations will show, these two types of information indeed have commonalities that contribute to the learning process in an MTL setting. In addition, the proposed MTL approach is favorable over other combination approaches such as training the models independently.



\subsection{Related Work}
\label{subsection:related-work}
There is a plethora of classification models that combine raw image information together with additional metadata, which generally fall under the umbrella of multi-modal classification. Notable examples are image and text classification,
see \citet{baltruvsaitis2018multimodal} and \citet{ramachandram2017deep} for related surveys on multi-modal learning. Our work is fundamentally different than multi-modal learning since in this those works the other modalities are part of the features of the model. That is, they serve as input to the model both at training and inference time. In our work we assume that the other modality, namely collaborative information, is not available at inference time, yet we aim to leverage its existence at training time by using it to guide the learning of the model. Thus, the model may benefit from this signal without relying on it for inference.

Several works have tried to harness the power of CF for various image understanding problems. The work of \citet{dabov2007image}, for example, applies a CF technique to improve image denoising. \citet{choi2015collaborative}, on the other hand, showed how to leverage collaborative information for better pose estimation results. However, the neighborhood of an image in these works is determined not by the interactions with it, but according to visual similarity.

There are also existing several works that aim to improve the image content understanding process using collaborative information as we do in our work \cite{ma2010bridging, xu2019stacked, liu2019sparse, sigurbjornsson2008flickr}. However, in these works the user interaction signal is used as an input to their image understanding models. Therefore, the collaborative information cannot be used to model \emph{new} images, which is the focus of our work. Our proposed method in contrast does not treat user interactions as a feature for the model, but only as means of guidance.

Another type of related work can be found in \citet{li2016weakly} and \citet{li2018deep}, which both aim to improve the training of content understanding models using user data. However, they treat the user interaction data as a type of content. That is, they omit the identities of the interacting users and do not apply collaborative filtering as we do. As a result of such an approach, if the user interaction is, for example, image \emph{tagging}, then their work is limited to enhancing image tagging processes. An earlier work described in \citet{sang2012user} retains the users' identities and its objective is to improve image tagging using interaction data. This is done by applying tensor factorization to model users, images and tags in the same joint latent space. However, this approach also binds the type of the interactions and the prediction task. In our approach, the prediction task is not tied to a specific type of user interactions.

Common to all discussed approaches based on user data is their need to process the raw image data for each user interaction. As the number of pixels $P$ in a typical image reaches to million ($10^6$) and the number $N$ of interactions in real-life datasets can be hundreds of millions ($10^8$), their running time is in $O(N\cdot P)=10^{14}$, which is computationally expensive or even infeasible for large real-life datasets. Differently from that, in our approach we first apply a CF method on the interaction data to create a compact representation for each item of dimensionality $f$, and we then train a content understanding model using those fixed-size vectors. Therefore, the running time is $O(f\cdot (P+N)) \ll O(N\cdot P)$. This is achieved since instead of taking a naive end-to-end solution, we adopt a two-phase solution, which first transforms the interaction data to a label set.

Finally, it is worth mentioning that the other direction, i.e., content understanding to enhance recommender systems, has shown striking success \cite{barkan2019cb2cf, he2016vbpr, sanchez2012social, shalom2019generative, eden2022investigating}. The relationship of these works to our current approach is however limited, as our goal is to improve the image understanding process and not item recommendations.

\section{Proposed Technical Approach}\label{sec:approach}

\subsection{Overview}

\begin{figure*}[btp]
    \centering
    \includegraphics[width=1.0\textwidth]{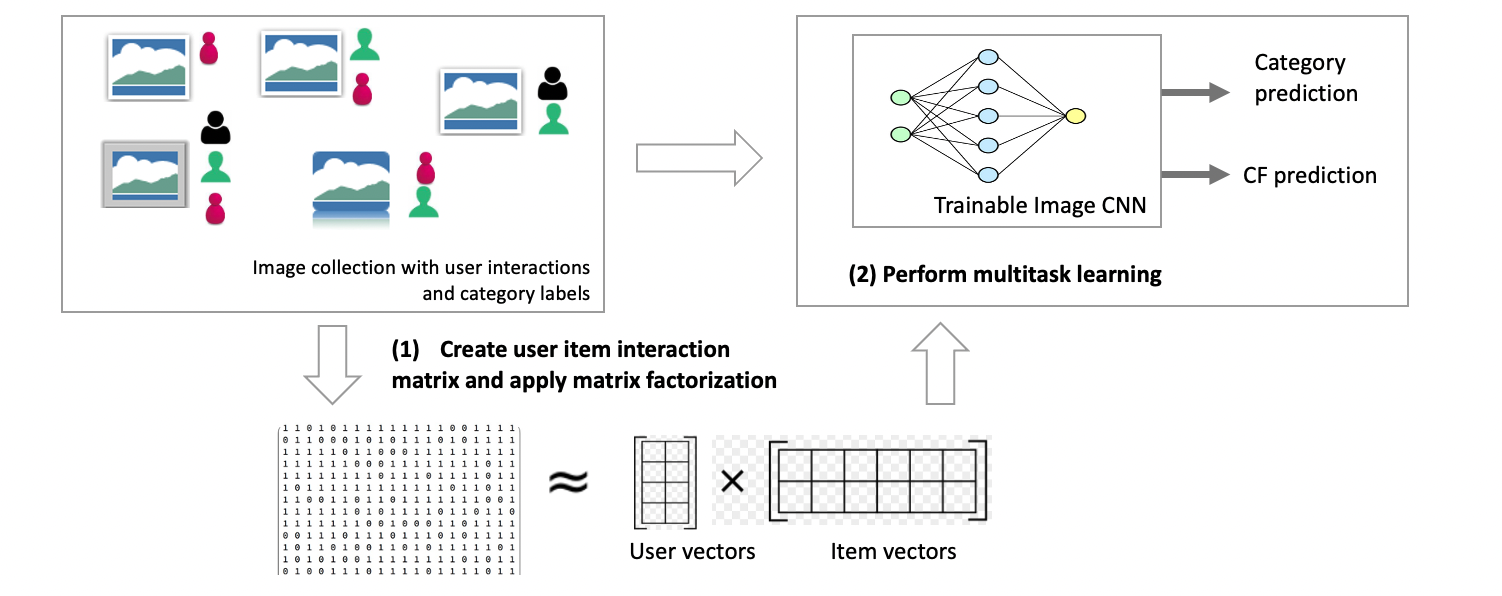}
    \caption{Overview of the proposed method.}
    \label{fig:method-overview}
\end{figure*}

Borrowing terminology from the recommender systems community, the term \emph{item} refers to an entity that users can interact with, and furthermore may have an image as side-information. In this paper, as the main focus is image understanding, we use the terms image and item mostly interchangeably.
In some cases the image can be the item itself, as in image sharing services (e.g., Pinterest) for example, where users interact with images.

Our proposed approach consists of two main phases, as illustrated in Figure~\ref{fig:method-overview}. The input to the learning task is a collection of images, where (i) each image has category labels assigned and (ii) for each image we are given a set of users who have interacted with it, e.g., through rating, tagging, or sharing. In the \emph{first phase} of our approach, a user-item interaction matrix is derived from the input data and a CF technique is applied to create fixed-size latent representations (vectors) of users and items. The purpose of this phase is solely to generate augmented labels for the classification problem. In the \emph{second phase}, multitask learning is applied, where the main task is to predict the categories of the images, and the auxiliary task is to reconstruct the latent item representations created in the first phase\footnote{The latent user vectors are not used in this process.}. As mentioned, we assume that the latent item vectors---dubbed \emph{CF vectors} from here on---encode category information that can help the training of the main task. Technically, instead of learning the image categorization from scratch, we rely on a pre-trained image model, which we then fine-tune with the given image data of the problem at hand.

Once the model is trained, we use it to predict categories of \emph{new images}, i.e., for images for which we do not have collaborative information. Typical important use cases could be pictures that users upload on a social media platform or pictures of newly added shop items that should be automatically categorized.

It is important to note that at this level, we consider our technical contribution as a \emph{framework} because we propose a general form of incorporating collaborative information into the image understanding process. In a given application scenario, the framework can then be instantiated in appropriate ways, using a suitable combination of matrix factorization technique, training set for pre-training, and deep learning architecture for image categorization.

\subsection{Technical Approach}


In the following, we describe our approach more formally. In this formalization, we use $i\in \{1,\ldots,I\}$ to index items. Table~\ref{tab:notations} summarizes the notations in this paper.
\begin{table}[tb]
\centering
\begin{tabular}{ cl }
 \hline
 Symbol & Explanation \\ \hline
 $x_i$ & Content of item $i$ \\
 $y_i$ & Category label of item $i$ (a binary vector)\\
 $\hat{y}_i$ & A predicted label of item $i$ (a real-valued vector)\\
 $U_i$ & Set of users who interacted with item $i$ \\
 $q_i$ & CF vector of item $i$ \\
 $\hat{q}_i$ & A predicted CF vector of item $i$ \\
 $\omega_i$ & Weight of CF vector of item $i$ \\
 \hline
\end{tabular}
\caption{Main symbols used in the paper.}
\label{tab:notations}
\end{table}

At training time, we assume each item $i$ is associated with an \emph{image} $x_i$ and with some \emph{label} $y_i$, a category in our use case. Furthermore, the item may be associated with a set $U_i$ of users who interacted with it.
At inference time, given an image to be modeled we assume it is \emph{not} associated with historical interaction data.

In many practical scenarios, the content understanding model must be invoked upon introducing the new item to the system, and therefore it has not gathered usage data (also known as \emph{cold item}, a ubiquitous problem in recommender systems \cite{sar2015data}). We emphasize that integrating collaborative filtering information at inference time has been proposed earlier, see Section~\ref{subsection:background}. These approaches assume the existence of such information for the images to be modeled. In contrast, this work focuses on situations where such information is not available. Integrating existing collaborative information into our framework is deferred to future work.



Therefore, the input for the model is a set of triplets $\{(x_i, y_i, U_i)\}$, where $x_i$ serves as the feature set and $y_i$ and $U_i$ provide guidance for training the model. We note that at training time some items $i$ might miss wither the label $y_i$ or the collaborative data $U_i$, which frames the problem as a semi-supervised one \cite{chapelle2009semi}. As we show in Section~\ref{sec:experiments}, when only a portion of the training data is labeled, the relative contribution of our approach is even higher.

\subsection{Deriving CF vectors} In the first phase, the collaborative data $U_i$ is transformed into a binary user-item interaction matrix $M$. 
A collaborative filtering technique is then used to construct a user embedding matrix $P$ and an item embedding matrix $Q$ of a given size $f$ (number of factors). We highlight that any CF method can be applied, as long as it represents items by a fixed-size vector.
After this process, we are given, for each item $i$, a corresponding latent CF vector $q_i$. Although these vectors are uninterpretable, we hypothesize that they hold valuable information for the classification task, as indeed is empirically shown in Section~\ref{sec:experiments}.

\subsection{CF-guided Image Categorization} In the second phase, the image understanding (categorization) process is guided by the CF vectors obtained in the previous phase.

In our multitask-learning approach, predicting the labels is the main task and the auxiliary task is based on the given CF vectors. After performing the first task, the input to the model is a set of triplets $\{(x_i, y_i, q_i)\}$, where $y_i$ and $q_i$ are the labels. The model has two outputs, corresponding to the main and auxiliary tasks. $\hat{y}_i$ and $\hat{q}_i$ are the predicted categories and CF vector, respectively.

\paragraph{Main task: Classification loss} Generally, for classification problems with $N$ classes, the label $y$ is an $N$-dimensional binary vector (which can support multi-label classification), where $y^n$ denotes the $n$th entry in the vector. The single-task categorization model processes the input and then outputs a prediction vector $\hat{y}$ that aims to minimize the binary cross-entropy \cite{ruby2020binary}. Since labels might be imbalanced, that is, some categories may appear much more often than others, we also apply class balancing as follows. Let $\eta_n$ denote the number of appearances of class $n$. Then the raw weight of this class is $\frac{1}{\eta_n}$. We normalize the weights such that the average weight is 1, that is: $\frac{1}{N}\sum_{n=1}^N \frac{1}{\eta_n} = 1$. Then the class-balanced binary cross-entropy is defined as:
\begin{equation}
\ell_{MAIN}(\hat{y}, y) = \sum_{n=1}^N \frac{1}{\eta_n} y^n\cdot \log \hat{y}^n + (1-y^n)\cdot \log (1-\hat{y}^n).
\end{equation}



\paragraph{Auxiliary task: CF-based loss}
Inspired by \citet{gou2021knowledge}, we propose a method that considers all of the information embedded in the CF vectors to reconstruct the CF vector using the content only. We therefore name this method \mtlreconstruct.
Let $L_1$ and $L_2$ denote the Manhattan and Euclidean distances between the CF vector $q$ and its reconstruction $\hat{q}$, respectively. The reconstruction loss (denoted as $l_{CF\_R}$) asks to minimize the difference between the vectors as follows\footnote{We experimented with several other metrics, like convex combinations and single distance loss. Yet, this one yielded the best results.}:
$$\ell_{CF\_R}(q_i, \hat{q}_i) = exp(L_1(q_i, \hat{q}_i) + L_2(q_i, \hat{q}_i))$$

\paragraph{Considering CF vector relevance weights.} 
Since CF methods do not converge to a global optimum, some items might be poorly represented and therefore the CF vectors induce weak supervision \cite{hoffmann2011knowledge}.
To cope with such ``noisy labels'' in our problem setting, where some CF vectors might be more informative than others, we introduce a per-instance confidence weighting method which affects the impact of the auxiliary task. This method may depend on the CF vector $q$ or even on the raw collaborative information $U$, annotated as $h(\cdot)=\omega$. To avoid shattering gradients, $\omega$ is upper bounded by some hyperparameter $\Omega$. The auxiliary loss function is then:
\begin{equation}
    \ell_{AUX}(\hat{q}, q, \omega) = min(\omega, \Omega) \cdot \ell_{CF\_R}(q, \hat{q})
\end{equation}

We explored three ways of weighting the CF vectors when training the model with the \mtlreconstruct approach.

\begin{itemize}
  \item \emph{Uniform confidence}: This approach neglects the possible variability in the fitness and information encoded in the CF vectors. Hence, $h$ is a nullary function which assigns the same weight to each CF vector.
  \item \emph{Interaction confidence}: The intuition behind this approach is as more users interacted with an item, then there is more collaborative information about it, and as a consequence we trust its produced CF vector more. Therefore, according to these considerations, there should be a positive correlation between the assigned weight of a CF vector and the number of interacting users. Let $|U_i|$ denote the number of interacting users with item $i$. Then\footnote{Another plausible alternative is linear correlation, however it was inferior to the proposed method in preliminary experiments.}, $h(U_i)=\sqrt{|U_i|}$.
  \item \emph{Loss-based confidence}: The utility of a CF vector for the category prediction task might not always be proportional to its amount of interacting users.
  We therefore suggest a novel method that explicitly captures the utility level of each individual CF vector according to its intrinsic traits. To this end, we train a category prediction model, called \emph{\cftolabel}, that takes as input only the CF vector (without the image itself). See Section ~\ref{sec:pre-study} for more details on this model.
We then approximate the fitness of a CF vector by its ability to predict the labels, measured by the mean reciprocal loss value on the positive labels. Formally, let $y$ be the actual label of a given item and $\hat{y}$ be the predicted label: $\cftolabel(q)=\hat{y}$. The mean loss on the positive labels is:
  $$\ell_{WEIGHT}(\hat{y}, y) = \frac{1}{\sum_n y^n} \sum_{n=1}^N y^n\cdot \log \hat{y}^n. $$
Hence, the weight of this CF vector is:
$$h(q_i, y_i) = \frac{1}{\ell_{WEIGHT}(\cftolabel(q_i), y_i)}.$$
Our experiments will show that this novel weighting method considerably contributes to the performance of the model. We also experimented with the mean across all labels, including the negative ones. However, this deteriorated results, as the performance on positive labels is more informative.
\end{itemize}

\paragraph{Weighting the tasks.}
Various ways of learning how to balance the weights of the tasks in multitask-learning were proposed in the literature \cite{chen2018gradnorm,guo2018dynamic,kendall2018multi}. However, since our problem involves only two tasks, we resort to a simple hyperparameter tuning strategy to find suitable weights. All in all, the final loss function is:
\begin{equation}
    \ell_{MTL}(\hat{y}, y,\hat{q},q,\omega) = \ell_{MAIN}(\hat{y}, y) + \alpha\cdot \ell_{AUX}(\hat{q},q,\omega),
\end{equation}
where $\alpha$ is a hyperparameter that controls the relative importance of the two tasks.
In all experiments, the optimal weight of the auxiliary task was non-negligible ($1<\alpha<2$), and our experiments clearly demonstrate the importance of the auxiliary task.
\color{black}

\section{Experimental Evaluation}\label{sec:experiments}
Next, we report the results of an in-depth experimental evaluation of our MTL-based image understanding approach. Specifically, we seek to answer the following research questions (RQs).
\begin{itemize}
\item (RQ1) To what extent does our method improve image understanding processes \emph{(a)} compared to models that do not use collaborative information and \emph{(b)} compared to alternative ways of incorporating collaborative information?
\item (RQ2) Given that missing labels are common in real-world scenarios, we ask: What is the relative gain of the proposed method in case of data paucity, i.e., when many labels are missing in the training data?
\end{itemize}

\subsection{Experiment Setup}
\subsubsection{Datasets}
\label{subsec:datasets}
To show the robustness of our approach we made experiments with four real-world datasets. These datasets differ in fundamental properties like domain and types of labels. Each of them contains item images, category labels, as well as user interaction data. Furthermore, they were selected such that they cover very popular domains: e-commerce, online content and social media.

Two of the datasets are from the e-commerce domain and were collected from Amazon.com~\cite{mcauley2015image}. The items in these two public datasets are from the area of ``Clothing'' and ``Toys'', where products are associated with their catalog images. The labels in these datasets correspond to the item categories and user interactions refer to purchases. The third dataset is a version of the widely-used public MovieLens 10M dataset \cite{harper2015movielens}, which includes movie posters. 
In this dataset, the movie genres are used as category labels and user ratings as interactions. Finally, as a fourth dataset we include a dataset from the social media domain. It is an anonymized version of the Pinterest image dataset used in \cite{zhong2015predicting}\footnote{\url{https://nms.kcl.ac.uk/netsys/datasets/social-curation/dataset.html}}. Here, the items are the images themselves; the labels correspond to user-provided tags and interactions are repin activity\footnote{\url{https://sproutsocial.com/glossary/repin/}}. The labels here are single-class, unlike the rest of the datasets, which have multiple labels per item.

For all datasets, the recorded user interactions, e.g., ratings, serve as collaborative signals. The main statistics of the datasets are provided in Table \ref{tab:dataset_statistics}. Observe that the datasets not only vary in terms of size and density, but also with respect to the average number of categories each item is assigned to (``Avg. pos. labels'').




\subsubsection{Pre-Trained Image Models}
We conducted 
preliminary experiments with different pre-trained models versus randomly initialized image models. These experiments confirmed that in all cases fine-tuning a pre-trained image model leads to substantially better classification results. 
In the rest of the paper, we therefore only report results that were obtained with pre-trained models.

\begin{table}[htb]
    \centering
    \begin{tabular}{l|cccc}
\toprule
Statistic & Pinterest & MovieLens & Clothing & Toys \\
\toprule
\#Images        & 27,080    & 10,677        & 23,033    & 11,924    \\
\#Interactions  & 265,252   & 10,000,054    & 278,677   & 167,597   \\
\#Users         & 170,978   & 68,532        & 39,387    & 19,412    \\
Sparsity        & 5.73e-5          & 1.37e-2          & 3.07e-4   & 7.24e-4      \\
\#Labels        & 32        & 19            & 199       & 54        \\
Avg. pos. labels & 1.0 & 2.03 & 3.22 & 1.28 \\
Gini index  & 0.48 &  0.51 & 0.74 & 0.58  \\
\bottomrule
\end{tabular} 
    \caption{Dataset statistics.}
    \label{tab:dataset_statistics}
\end{table}


Over the last years, a variety of deep learning architectures were proposed that can be used for image categorization, as discussed in Section~\ref{sec:background}. One goal of our experiments is to demonstrate the robustness of our MTL-based approach with respect to the underlying image model, i.e., that the effectiveness of our approach is not tied to a particular image model and may contribute to new models to come in the future. Therefore, we considered the following three state-of-the-art architectures in our experiments: ResNet18 \cite{he2016deep}, MobileNet \cite{howard2017mobilenets}, and RegNet \cite{radosavovic2020designing}. To avoid any bias, we randomly assigned each of them to one of the datasets, ending up with ResNet18 for the Amazon datasets, RegNet for Pinterest, and MobileNet for MovieLens. All models were pre-trained on the ImageNet set \cite{krizhevsky2012imagenet}.


\subsubsection{CF Models}\label{sssec:cf_models}
Generally, we hypothesize that the CF vectors carry meaningful information about the items, and that this information can be advantageous for our problem, which we validate in Section~\ref{sec:pre-study}. To incorporate a well fitted CF model in our approach, we seek a method that performs well for the given datasets. A large variety of CF models that use item embeddings were proposed over the last fifteen years, and we experimented with several landmark CF methods. Among other methods, we considered
Bayesian Personalized Ranking (BPR) as a more traditional model \cite{rendle2009bpr} and Variational Autoencoders (VAE) \cite{liang2018variational}, which is a state-of-the-art technique\footnote{We relied on the implementations provided in the Cornac framework (\url{https://github.com/PreferredAI/cornac})}.
Motivated by \cite{press2016using}, the last layer of the decoder in the autoencoder architecture serves as the item embedding matrix $Q$.
Moreover, we supplemented the original VAE implementation with bias terms as this led to even better results. Note that we also tried an autoencoder-based recommender based on the VAE model proposed in \cite{ouyang2014autoencoder}. This however did not lead to competitive results.

In Table~\ref{tab:recommender}, we report the results of BPR and VAE in terms of the AUC metric for the four datasets after careful fine-tuning. For comparison purposes, we also include the results obtained by a non-personalized recommender that always suggests the most popular items to users. Not surprisingly, the results show that VAE consistently works better than BPR across all datasets. Therefore, we use VAE as the CF model for all experiments reported in this paper.
Remember that we do not rule out that other CF models may exist that lead to even better recommendation results for individual datasets, and we iterate that finding the best existing CF model is not the main focus of our present research. Nonetheless, we acknowledge that future advances in CF models may be beneficial for our MTL-based image categorization as well.


\begin{table}[tb]
    \centering
\begin{tabular}{l|cccc}
\toprule
Method        & Pinterest & MovieLens & Clothing & Toys\\
\midrule
Most popular    & 0.47 & 0.94 & 0.63 & 0.65  \\
BPR             & 0.56 & 0.97 & 0.66 & 0.69 \\
VAE             & 0.96 & 0.97 & 0.75 & 0.82  \\
\bottomrule
 \end{tabular}
    \caption{Recommendation performance of CF models.}
    \label{tab:recommender}
    \vspace{-20pt}
\end{table}

\subsubsection{Evaluation Metric}
To measure classification performance apply the widely-used Mean Average Precision (mAP) metric \cite{hossin2015review}, arguably one of the most adopted evaluation metrics for multi-label classification (e.g., \cite{Xu2007AdaRankAB,revaud2019learning,Zha2015ExploitingIC,kaufman2019balancing}). mAP is computed as:
$$\frac{1}{N} \sum_{c=1}^N AP(c) $$
where $N$ is the number of classes and $AP(c)$ denotes the average precision in class $c$. To compute statistical significance we use bootstrapping~\cite{park2011bootstrap} and paired sample $t$-tests~\cite{weisstein2001student}.




\subsubsection{Implementation \& Hyperparameter Tuning}
All of the implementation for our experiments was done in Python, using the PyTorch framework. We share all material online for reproducibility\footnote{\url{https://github.com/anonymous1e6/cactus}}.
Model selection and hyperparameter tuning (e.g., learning rate, regularization and number of hidden layers and sizes) and test were generally done in a systematic way on a held-out set, evenly divided into evaluation and test sets.
The held-out set for the classification models was created by randomly selecting 30\% of the items. We emphasize that the collaborative information of these items was omitted as well, to compute the performance on images with no prior user interactions.

The held-out dataset for the recommender systems was created by randomly selecting 20\% of the interactions and AUC \cite{lu2012recommender} was used as the evaluation metric. This metric was chosen instead of top-$k$ evaluation metrics like precision@$k$ or nDCG@$k$ because all test items directly contribute to the final measure and not just those (possibly few) items that made it to a top-$k$ list. This is an important trait because we care about the fitness of all items, as each of them contributes to the auxiliary task in the MTL setting.

\subsection{Pre-Study: CF-Based Category Prediction}
\label{sec:pre-study}
Before we discuss the results of the evaluation of our approach to collaborative image understanding in Section~\ref{ssec:results}, we report the outcomes of a pre-study in which we aimed to validate the value of collaborative information for category prediction.

To this end, we conducted the following experiment in which the image category prediction task is solely based on the available interaction data, i.e., the matrix describing user-item interactions. As a first step in the experiment, we trained a collaborative filtering model on the interaction matrix to obtain the needed latent item vectors.
Specifically, we carefully tuned
BPR and VAE CF models
on the four evaluated datasets. The resulting fixed-size\footnote{A latent vector size of 64 worked best for all datasets; smaller or larger values led to degraded recommendation performance.} latent item vectors  $q_i$ then serve as the input to a neural network, which predicts the corresponding image category label $y_i$. We name these models \emph{\cftolabel}.
The model used in our experiments consists of a two-layered MLP with ReLU as the activation function and Binary Cross-Entropy per each class as the loss function\footnote{Other model architectures are possible as well. Remember that the main here goal is to demonstrate the general value of interaction data, but not necessarily to find the best possible architecture.}.

The results of this experiment are shown in Table~\ref{tab:cf_based}, where we report the image categorization performance in terms of mAP. As a reference point to gauge the effectiveness of the trained models, we provide the mAP results that we obtain when making predictions solely based on the popularity of the categories in the different datasets, i.e., when we always predict the most popular category.

\begin{table}[tb]
    \centering
    \begin{tabular}{c|cccc}
\toprule
Approach                &  Pinterest        & MovieLens         & Clothing          & Toys \\
\midrule
Popularity based        & 0.031             & 0.107             & 0.016             & 0.024 \\
\cftolabel \ BPR        & 0.075             & 0.465             & 0.021             & 0.074 \\    
\cftolabel \ VAE        & 0.266             & 0.528             & 0.124	            & 0.241 \\    
\bottomrule
\end{tabular}
    \caption{Performance results for \emph{\cftolabel} (mAP).}
    \label{tab:cf_based}
    \vspace{-20pt}

\end{table}

The results clearly show that even though the CF item vectors were trained oblivious to the categories, they indeed contain valuable information for the category prediction task. The mAP values obtained with the \emph{\cftolabel} variants are up to ten times higher than those of the popularity-based approach. Aligned with our assumption stated in Section \ref{sssec:cf_models}, we furthermore obtained evidence that the VAE CF model, which outperformed BPR on the recommendation task, also outperformed it on this task and therefore should be used in our experiments.
On an \emph{absolute} scale, the highest mAP values are obtained for the MovieLens dataset. This is not surprising, given that it has the lowest number of categories to predict and at the same time the highest user-item interaction density across all datasets. In terms of the \emph{relative} improvement over the baseline, the gains obtained by \emph{\cftolabel} are however the lowest across the datasets, i.e., a five-time improvement.

Let us emphasize here that the mAP results reported in Table~\ref{tab:cf_based} must not be compared on an absolute scale with the mAP values reported next in Section~\ref{ssec:results}. In the pre-study reported here, the CF-based image category predictions were based on knowledge about the observed user interactions for a given image to categorize. Our MTL-based approach described in Section~\ref{sec:approach}, however, is designed for situations in which no collaborative information is yet available, i.e., for new images, and where the collaborative information is merely used to guide the learning process of the main task. Remember here that we use the \emph{\cftolabel} method to inform the weighting of the CF vectors in the loss-based approach described in Section~\ref{sec:approach}.

\subsection{Baselines}
To answer RQ1 on the effectiveness of \mtlreconstruct and its variations, we, first of all, benchmark its performance against the performance that is achieved with the fine-tuned image model alone, i.e., with a model that does not use collaborative information. We refer to this baseline as \imageonly in the rest of the paper.

Regarding other possible baselines, we are not aware of any other work which used collaborative information in combination with image models to categorize \emph{new} images, that is, images for which no collaborative information is available yet. Therefore, we designed two alternative ways of how one could intuitively approach the problem: one of them based on multitask learning, and one based on sequential fine-tuning of the auxiliary task and the main task.

\subsubsection{Contrastive-Loss Approach}
A possible advantage of the proposed \mtlreconstruct approach is its per-dimension learning guidance by utilizing the entire information in the CF vectors. However, reconstructing the entire CF vector means simultaneously predicting tens or even hundreds of latent features. This might be an overwhelming task, and consequently might introduce noise to the training process.
Hence, we propose an alternative approach and thereby an additional baseline for our experiments, that resorts to supervised contrastive loss \cite{khosla2020supervised}. Namely, at each epoch it generates a list of triplets, where each item participates once as an \emph{anchor}. For any anchor item $a$ (associated with CF vector $q_a$), it first chooses a \emph{positive} item $p$ such that its associated CF vector $q_p$ is the most similar to $q_a$ in terms of the $L_2$ distance.
Then it chooses at random a \emph{negative} item $n$. The auxiliary task is then to distinguish between the positive and the negative items solely based on the predicted CF vectors. More formally, the task is to minimize the triplet loss \cite{schroff2015facenet}:
$$\ell_{cont}(q_a, q_p, q_n) = max\{L_2(q_a,q_p) - L_2(q_a,q_n) + \tau, 0\}$$
where $\tau$ is a hyperparameter, which serves as a positive margin. We refer to this baseline in our experiments as \contrastiveloss.

Compared to the \mtlreconstruct method, this approach induces an easier learning task, namely a binary classification task rather than a multidimensional regression task. However, a substantial amount of information is lost, as it reduces the entire CF vector into a single scalar that represents the distance to the anchor.

\subsubsection{Sequential Approach}
A possible complication with multitask learning in general is that the auxiliary task might overshadow the main task. In such situations, the model tends to excel at the auxiliary task at the expense of the primary one. Therefore, we explore fine-tuning \cite{Too2019ACS,Campos2017FromPT} as an alternative way to leverage the available collaborative information for the main task. In this approach, the model in the first step is optimized only for the auxiliary task. Then, a fine-tuning phase follows where parameters are optimized for the primary task only. We refer to this method as \finetuning in the rest of the paper. We note that this baseline cannot benefit from a joint learning process as is the case for MTL.

\subsection{Results}\label{ssec:results}

\subsubsection{RQ1: Effectiveness of \mtlreconstruct}
Table~\ref{tab:main-result-classification} shows the main results of our experimental evaluation with respect to RQ1.
Considering the last row of Table~\ref{tab:main-result-classification}, we observe that our method \mtlreconstruct with the loss-based weighting scheme leads to consistent and solid performance improvements from 2.2\% up to 9.5\% over the \imageonly baseline across the different datasets.
The greatest improvements are observed for the Clothing dataset, which also has the highest number of categories.

The smallest improvements were achieved for the MovieLens dataset. This is consistent with the findings from Section~\ref{sec:pre-study}, where we also found the smallest relative gain from the latent item vectors with the \emph{\cftolabel} method for the MovieLens data. Using other weighting schemes for \mtlreconstruct was similarly beneficial except for one case for the MovieLens dataset, where the interaction-based loss did not lead to any improvements.

While \mtlreconstruct is effective for all datasets, the extent of the gain varies across the datasets. Various aspects may contribute to these differences, including dataset characteristics (e.g., class distributions, density of user interactions), aspects related to the item pictures (e.g., movie posters are artistic whereas e-commerce items are not), or the underlying idiosyncrasies of each domain.

\begin{figure*}[tb]
    \centering
    \includegraphics[width=1.0\textwidth]{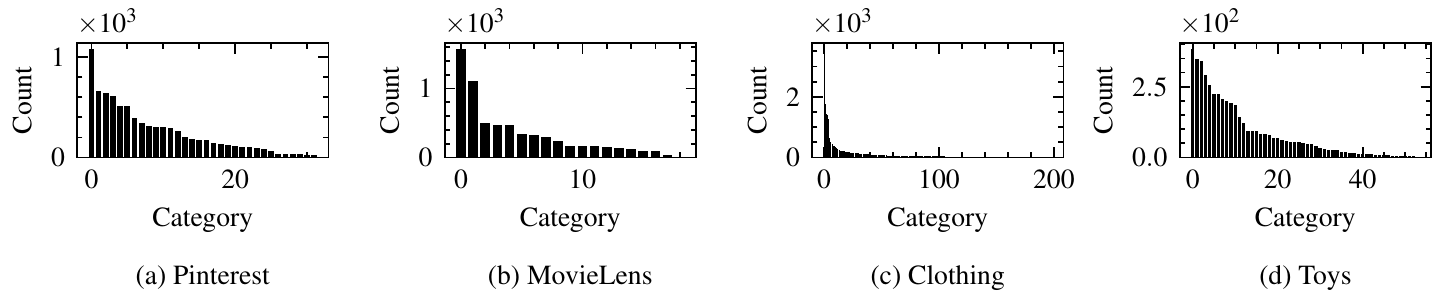}
    \vspace{-0.5cm}
    \caption{Samples count for each class.}
    \label{fig:label-count}
\end{figure*}

With respect to dataset characteristics, Figure~\ref{fig:label-count} illustrates the number of items per category for all four datasets. Let us consider Figure~\ref{fig:label-count}(c) and Figure~\ref{fig:label-count}(d), which show the distribution for the Clothing and Toys datasets. Note that even though both datasets are from the e-commerce domain and even from the same platform, we can observe a clear difference in terms of the shape of the long tail distribution.
In particular, the concentration on the ``short head'' is much more pronounced for the Clothing dataset, i.e., a large fraction of the items are assigned to a small set of broad (or popular) categories. The Gini index for these datasets is given in~\autoref{tab:dataset_statistics} and it supports these visual observations\footnote{Higher Gini values indicate a strong concentration of the samples.}.
Our experiments therefore indicate that the use of collaborative information may be particularly helpful to improve the image categorization process when there is a long tail of rather rare categories.
A deeper analysis of factors that influence the effectiveness of \mtlreconstruct is part of our future work.

We found no consistent performance gains for the baseline \contrastiveloss. While it worked very well for the Clothing dataset, for the Toys dataset, it almost had no effect; and it led to a performance degradation compared to the \imageonly model for the MovieLens and Pinterest data. Still, we conclude that the \contrastiveloss method might be a candidate to explore in certain practical settings, given that there are indications that it may work well in some domains.

Looking at the \finetuning approach, we find that it did not help to improve the results compared to the \imageonly baselines. In contrast, we actually consistently observed a small degradation when combining collaborative information with the image-based models in this way. This highlights the importance and value of combining collaborative information with image-based models using a shared representation.

\paragraph{Weighting the CF vectors} Comparing the naive weighting method \mtlreconstruct$_{\text{\!\emph{interaction}}}$ to the method that does not weigh the CF vectors \mtlreconstruct$_{\text{\!\emph{uniform}}}$, we do not find a clear winner. However, our novel weighting method \mtlreconstruct$_{\text{\!\emph{loss}}}$ consistently outperforms the rest of the methods. This proves the need to deeply inspect the CF vectors to assess their weights.


\begin{table*}[tb]
    \centering
    \begin{tabular}{r|cccc}
\toprule
Approach                                        &  Pinterest        & MovieLens         & Clothing          & Toys \\
\midrule
\imageonly                                      & 0.345             & 0.329             & 0.387             & 0.412 \\ \midrule
\finetuning                                     & 0.344 (-0.3\%)    & 0.308 (-6.5\%*)  & 0.352 (-9.0\%*)  & 0.41 (-0.4\%) \\ 
\contrastiveloss                                & 0.34 (-1.4\%*)   & 0.325 (-1.3\%*)   & 0.416 (+7.6\%*)    & 0.411 (-0.2\%)  \\ \midrule
\mtlreconstruct$_{\text{\!\emph{uniform}}}$     & 0.36 (+4.3\%*)   & 0.336 (+2.0\%*)  & 0.391 (+1.1\%*)    & \textbf{0.424 (+3.0\%*)}  \\ 
\mtlreconstruct$_{\text{\!\emph{interaction}}}$ & 0.361 (+4.6\%*)  & 0.329 (0.0\%)     & 0.392 (+1.4\%*)    & 0.417 (+1.3\%) \\           
\mtlreconstruct$_{\text{\!\emph{loss}}}$        & \textbf{0.363 (+5.2\%*)}  & \textbf{0.338 (+2.6\%*)}  & \textbf{0.422 (+9.1\%*)}    & \textbf{0.424 (+3.0\%*)}   \\
\bottomrule
\end{tabular}

    \caption{Image categorization performance results for four datasets (mAP). The best results for each dataset are printed in bold font. Increases over the \imageonly baseline that were found to be statistically significant are marked with * ($p<0.01$).}
    \label{tab:main-result-classification}
\end{table*}

\subsubsection{RQ2: Gains in situations of data paucity.}
A certain lack of labels in a given target domain is a common problem in many practical applications, given that this task is often manual and thus time-consuming, see also Section~\ref{sec:background}. Therefore, we analyzed the effectiveness of our proposed method in such situations of data paucity through a series of additional experiments.
Specifically, we measured the performance of our method compared to the baseline model \imageonly after randomly removing a certain fraction of the labels from the training data.

\begin{figure*}[h!tb]
    \includegraphics[width=0.95\textwidth]{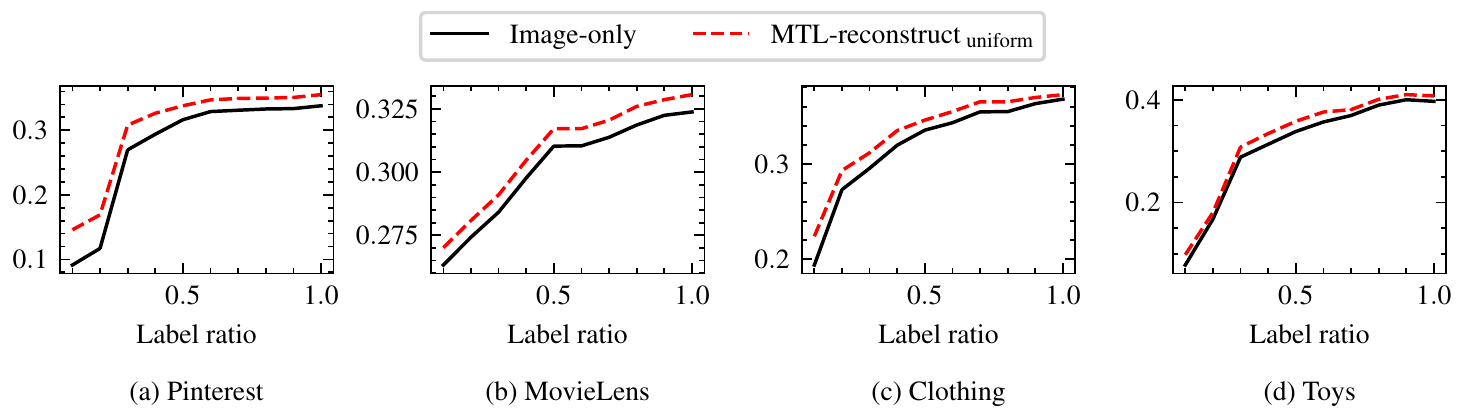}
    \vspace{-0.2cm}
    \caption{mAP values for \imageonly and \mtlreconstruct$_{\text{\!\emph{uniform}}}$ for different levels of missing labels.}
    \label{fig:map}
\end{figure*}

\begin{figure*}[h!tb]
    \includegraphics[width=0.95\textwidth]{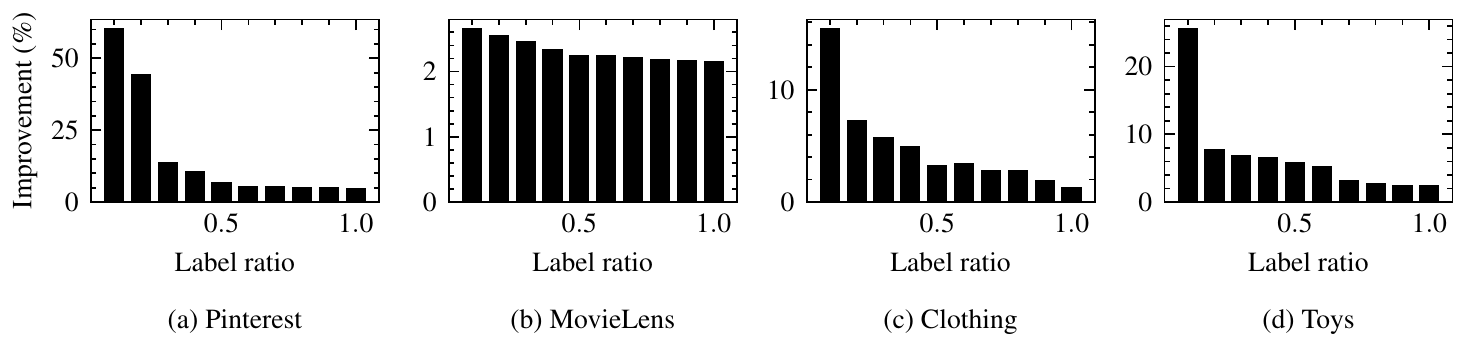}
    \vspace{-0.2cm}
    \caption{Relative mAP improvement of \mtlreconstruct$_{\text{\!\emph{uniform}}}$ over \imageonly for different levels of missing labels.}
    \label{fig:map_improvement}
\end{figure*}

Figure~\ref{fig:map} shows the results of this measurement for the compared methods in terms of the mAP metric. This figure is supplemented with confidence intervals, implemented using bootstrapping~\cite{park2011bootstrap}. Note that in the Pinterest dataset the area of the confidence interval is almost imperceptible, due to low variance in the results. Figure~\ref{fig:map_improvement} shows the relative improvement of \mtlreconstruct over the \imageonly baseline approach for the four datasets. On the horizontal axis, we show the ``Label ratio'', which denotes the percentage of labels that were retained in the training data. A value of 0.1 for the label ratio thus means that 90\% of the labels were discarded for this measurement. 

Note that we used the \mtlreconstruct$_{\text{\!\emph{uniform}}}$ variant for these experiments (and not the best-performing \mtlreconstruct$_{\text{\!\emph{loss}}}$) because only the uniform weighting scheme described in Section~\ref{sec:approach} can be applied due to the missing labels in the training data.

Looking at the results shown in Figure~\ref{fig:map} and Figure~\ref{fig:map_improvement}, we find that \mtlreconstruct is consistently better or at least performing equally well as the \imageonly method for all datasets and label ratio levels. For all datasets except MovieLens, we furthermore observe the pattern that the gains of using CF information in our approach are stronger when there are more missing labels. In these cases, considering the collaborative information seems particularly helpful and consistent with the intuition that the value of CF information increases when other information is lacking. For the dense MovieLens dataset, such a phenomenon is not observed, but the accuracy gains are again rather small on average (around 2\%) for this dataset in general.

Note that for the figure showing relative improvements (Figure~\ref{fig:map_improvement}), the gains appear very small or non-existent in some cases due to the scale of the y-axis. In fact, the improvements indicated by the smaller bars are in the range of a few percent as shown in Table~\ref{tab:main-result-classification}.


\section{Conclusions and Outlook}\label{sec:discussion}
We showed how collaborative information, which is often available in practical applications, can support image understanding processes. Our technical proposal is built upon principles of multitask learning and leads to consistent performance improvements over different baselines on four datasets. Considering that nowadays improvements over state-of-the-art models are rather marginal, our improvements are even more striking. Differently from earlier approaches, our work focuses on the categorization of new images for which also no collaborative information is yet available. Overall, our results however show that relevant signals for image categorization may be contained in the collaborative information that already exists for other images.

In terms of ongoing and future work, one important direction is to further explore the relative importance of the CF model used in the MTL framework and if alternative CF models help to further improve the overall performance.
In addition, we would like to apply our \mtlreconstruct$_{\text{\!\emph{loss}}}$ variant to the missing labels case instead of \mtlreconstruct$_{\text{\!\emph{uniform}}}$ we used in this work.
This can be done by estimating the prediction uncertainty as done in recent works (e.g., \cite{bibas2021single,sastry2020detecting,gal2016dropout,bibas2019deep,bibas2021distribution}).


There are two interesting directions to expand the general contribution of our proposed framework. First, while classification problems are arguably the most prominent problems in content understanding, it would be very useful to assess the usefulness of our proposed framework also for regression problems. Second, it is important to investigate if the proposed approach to collaborative content understanding is viable and effective also for other domains, in particular for text documents or multimedia content.

Going beyond such additional technical analyses and explorations, we anticipate future extensions to our work. Two natural extensions of key importance are as follows. One intuitive extension is to enable improved inference for items with existing collaborative information. Another worthy research approach is to achieve mutual reinforcement between collaborative filtering and content understanding by leveraging end-to-end learning. A plausible approach is to simultaneously train the content understanding model and the CF model, 
which allows to transfer valuable information between them.
\FloatBarrier

\balance

\bibliographystyle{ACM-Reference-Format}
\bibliography{sample-base}

\end{document}